\documentclass[letterpaper]{article} 
\usepackage{aaai19}  
\usepackage{times}  
\usepackage{helvet}  
\usepackage{courier}  
\usepackage{url}  
\usepackage{microtype}
\usepackage{graphicx}
\usepackage{textcomp}
\usepackage{booktabs} 

\usepackage{latexsym}

\usepackage{bbm}
\usepackage{amsmath}
\usepackage{multirow}

\usepackage{xspace}
\usepackage{enumitem}
\usepackage{subcaption}
\usepackage[all]{nowidow}

\usepackage{color}
\usepackage{algorithm}
\usepackage[noend]{algpseudocode}

\nocopyright

\DeclareMathOperator*{\argmax}{argmax}

\begin{document}
\title{Robust Text-to-SQL Generation with Execution-Guided Decoding}
\author{
    Chenglong Wang,\textsuperscript{1}\thanks{Equal contribution. Work done during two respective internships at Microsoft Research.}~
    Kedar Tatwawadi,\textsuperscript{2}\footnotemark[1]~
    Marc Brockschmidt,\textsuperscript{3}~
    Po-Sen Huang,\textsuperscript{3}\thanks{Work done while at Microsoft Research.} \\
    {\bf \Large Yi Mao,\textsuperscript{3}~
    Oleksandr Polozov,\textsuperscript{3}~
    Rishabh Singh\textsuperscript{4}\footnotemark[2]} \\
    \textsuperscript{1}{University of Washington} \quad
    \textsuperscript{2}{Stanford University} \quad
    \textsuperscript{3}{Microsoft Research} \quad
    \textsuperscript{4}{Google Brain} \\
    \texttt{clwang@cs.washington.edu \quad kedart@stanford.edu} \\
    \texttt{\{mabrocks,pshuang,maoyi,polozov\}@microsoft.com \quad rising@google.com}
}
\maketitle
\newcommand{\fix}{\marginpar{FIX}}
\newcommand{\new}{\marginpar{NEW}}

\newcommand{\code}[1]{\texttt{#1}}

\newif\ifcomments
\commentstrue

\newcommand{\CWComment}[1]{\ifcomments\textcolor{blue}{\footnotesize CW: #1}\fi}
\newcommand{\CWMargin}[1]{\marginpar{\CWComment{#1}}}
\newcommand{\MBComment}[1]{\ifcomments\textcolor{cyan}{\footnotesize MB: #1}\fi}
\newcommand{\MBMargin}[1]{\marginpar{\MBComment{#1}}}
\newcommand{\RSComment}[1]{\ifcomments\textcolor{red}{\footnotesize RS: #1}\fi}
\newcommand{\RSMargin}[1]{\marginpar{\RSComment{#1}}}
\newcommand{\APComment}[1]{\ifcomments\textcolor{magenta}{\footnotesize AP: #1}\fi}

\newcommand{\size}[1]{\ensuremath{\lvert #1 \rvert}}
\newcommand{\intoks}[0]{\ensuremath{\mathit{X}}\xspace}
\newcommand{\intok}[0]{\ensuremath{\mathit{x}}\xspace}
\newcommand{\intokslen}[0]{\ensuremath{|\intoks|}\xspace}
\newcommand{\query}[0]{\ensuremath{Q}\xspace}
\newcommand{\querytag}[0]{\ensuremath{q}\xspace}
\newcommand{\querylen}[0]{\ensuremath{\query}}
\newcommand{\columns}[0]{\ensuremath{C}\xspace}
\newcommand{\columnstag}[0]{\ensuremath{c}\xspace}
\newcommand{\columnsnum}[0]{\ensuremath{\columns}}

\newcommand{\fwdtag}[0]{\ensuremath{\mathit{fw}}}
\newcommand{\bwdtag}[0]{\ensuremath{\mathit{bw}}}

\newcommand{\encodertag}[0]{\ensuremath{e}}
\newcommand{\embedding}[0]{\ensuremath{\mathit{emb}}}
\newcommand{\encemb}[0]{\ensuremath{\embedding_{\encodertag}}\xspace}
\newcommand{\enccellfwd}[0]{\ensuremath{\mathcal{C}_{\encodertag,\fwdtag}}}
\newcommand{\enccellbwd}[0]{\ensuremath{\mathcal{C}_{\encodertag,\bwdtag}}}
\newcommand{\encouts}[0]{\ensuremath{\mathit{O}_{\encodertag}}\xspace}
\newcommand{\encoutsym}[0]{\ensuremath{\mathit{o}}}
\newcommand{\encout}[1]{\ensuremath{\encoutsym^{(#1)}_{\encodertag}}}
\newcommand{\encoutfwd}[1]{\ensuremath{\encoutsym^{(#1)}_{\encodertag,\fwdtag}}}
\newcommand{\encoutbwd}[1]{\ensuremath{\encoutsym^{(#1)}_{\encodertag,\bwdtag}}}
\newcommand{\enchidsym}[0]{\ensuremath{\mathit{h}}}
\newcommand{\enchid}[1]{\ensuremath{\enchidsym^{(#1)}_{\encodertag}}}
\newcommand{\enchidfwd}[1]{\ensuremath{\enchidsym^{(#1)}_{\encodertag,\fwdtag}}}
\newcommand{\enchidbwd}[1]{\ensuremath{\enchidsym^{(#1)}_{\encodertag,\bwdtag}}}

\newcommand{\vocab}[0]{\ensuremath{\mathcal{V}}}
\newcommand{\voctype}[0]{\ensuremath{\tau_{\vocab}}\xspace}
\newcommand{\coltype}[0]{\ensuremath{\tau_{\columns}}\xspace}
\newcommand{\consttype}[0]{\ensuremath{\tau_{\query}}\xspace}

\newcommand{\decodertag}[0]{\ensuremath{d}}
\newcommand{\decemb}[0]{\ensuremath{\embedding_{\decodertag}}\xspace}
\newcommand{\deccell}[0]{\ensuremath{\mathcal{C}_{\decodertag}}}
\newcommand{\decouts}[0]{\ensuremath{\mathit{O}_{\decodertag}}\xspace}
\newcommand{\decinputsym}[0]{\ensuremath{\mathit{i}}}
\newcommand{\decoutsym}[0]{\ensuremath{\mathit{o}}}
\newcommand{\decinput}[1]{\ensuremath{\decinputsym^{(#1)}_{\decodertag}}}
\newcommand{\decout}[1]{\ensuremath{\decoutsym^{(#1)}_{\decodertag}}}
\newcommand{\dechidsym}[0]{\ensuremath{\mathit{h}}}
\newcommand{\dechid}[1]{\ensuremath{\dechidsym^{(#1)}_{\decodertag}}}

\newcommand{\decattenergy}[2]{\ensuremath{\mathit{u}^{(#1,#2)}}}
\newcommand{\decattvector}[1]{\ensuremath{\alpha^{(#1)}}}
\newcommand{\decattoutweight}[0]{\ensuremath{W_o}}
\newcommand{\decatthidweight}[0]{\ensuremath{W_h}}
\newcommand{\decattbias}[0]{\ensuremath{v}}

\newcommand{\decattVenerg}[2]{\ensuremath{\mathit{u}_{\vocab}^{(#1,#2)}}}
\newcommand{\decattVvector}[1]{\ensuremath{\alpha_{\vocab}^{(#1)}}}
\newcommand{\decattVoutweight}[0]{\ensuremath{W_{\vocab}^o}}
\newcommand{\decattVhidweight}[0]{\ensuremath{W_{\vocab}^h}}
\newcommand{\decattVbias}[0]{\ensuremath{b_{\vocab}}}

\newcommand{\decVweight}[0]{\ensuremath{W_{\vocab}}}
\newcommand{\decVbias}[0]{\ensuremath{b_{\vocab}}}

\newcommand{\decattCenerg}[2]{\ensuremath{\mathit{u}_{\columns}^{({#1},{#2})}}}
\newcommand{\decattCvector}[1]{\ensuremath{\alpha_{\columns}^{(#1)}}}
\newcommand{\decattCoutweight}[0]{\ensuremath{W_{\columns}^o}}
\newcommand{\decattChidweight}[0]{\ensuremath{W_{\columns}^h}}
\newcommand{\decattCbias}[0]{\ensuremath{b_{\columns}}}

\newcommand{\outtoks}[0]{\ensuremath{\mathit{O}}\xspace}
\newcommand{\outtok}[0]{\ensuremath{\mathit{o}}\xspace}

\newcommand{\progtoks}[0]{\ensuremath{\mathit{Y}}\xspace}
\newcommand{\progtok}[0]{\ensuremath{\mathit{y}}\xspace}
\newcommand{\lossV}[0]{\ensuremath{\mathit{loss}_{\vocab}}\xspace}
\newcommand{\lossCP}[0]{\ensuremath{\mathit{loss}_{\columns}^{\text{pntr}}}\xspace}
\newcommand{\lossCV}[0]{\ensuremath{\mathit{loss}_{\columns}^{\text{val}}}\xspace}


\newcommand{\rCH}[1]{Chapter~\ref{#1}}
\newcommand{\rSC}[1]{Section~\ref{#1}}
\newcommand{\rF}[1]{Figure~\ref{#1}}
\newcommand{\rD}[1]{Def.~\ref{#1}}
\newcommand{\rL}[1]{Lemma~\ref{#1}}
\newcommand{\rT}[1]{Thm.~\ref{#1}}
\newcommand{\rA}[1]{Alg.~\ref{#1}}
\newcommand{\rE}[1]{Ex.~\ref{#1}}
\newcommand{\rC}[1]{Cor.~\ref{#1}}
\newcommand{\rTab}[1]{Table~\ref{#1}}
\newcommand{\rComment}[1]{comment~\ref{#1}}
\newcommand{\rEq}[1]{\ensuremath{(\ref{#1})}}

\newcommand{\ignore}[1]{}
\long\def\ignorethis#1{}

\newcommand{\citet}[1]{\citeauthor{#1}~\shortcite{#1}}

\begin{abstract}
We consider the problem of neural semantic parsing, which translates natural language questions into {\em executable} SQL queries.
We introduce a new mechanism, {\em execution guidance}, to leverage the semantics of SQL. It detects and excludes faulty programs during the decoding procedure by conditioning on the execution of partially generated program. The mechanism can be used with any autoregressive generative model, which we demonstrate on four state-of-the-art recurrent or template-based semantic parsing models.
We demonstrate that execution guidance universally improves model performance on various text-to-SQL datasets with different scales and query complexity: WikiSQL, ATIS, and GeoQuery.
As a result, we achieve new state-of-the-art execution accuracy of 83.8\% on WikiSQL.


\end{abstract}


\section{Introduction}

\label{sec:intro}


Recent large-scale digitization of record keeping has resulted in
vast databases that encapsulate much of an organization's
knowledge.
However, querying these databases usually requires users to understand
specialized tools such as SQL, restricting access to this knowledge to select few. Thus, one aspiration of natural language processing is to translate natural
language questions into formal queries that can be executed automatically and
efficiently on a database, yielding natural interfaces to so-far inaccessible
repositories of knowledge for end users. Developing effective semantic parsers to translate natural language questions into logical programs has been a long-standing goal~\cite{DBLP:conf/acl/Poon13,Zettlemoyer:2005:LMS,DBLP:conf/acl/PasupatL15,nalix,nlyze}. In this work, we focus on the semantic parsing task of translating natural language queries into executable SQL programs.

As in many other research areas, recently introduced deep learning approaches have been very successful in this task.
A first generation of these approaches (e.g. by \citet{IyyerSQA2016}) has focused on adapting tools from (neural) machine translation, such as deep sequence-to-sequence (seq2seq) architectures with attention and copying mechanisms.
While often effective, such approaches commonly fail at generating \emph{syntactically valid} queries, and thus more recent work has shifted towards using grammar-based sequence-to-tree (seq2tree) models~\cite{krishnamurthy2017neural,DBLP_journals/corr/YinN17,DBLP_journals/corr/RabinovichSK17,sqlnet,coarse2fine}.
In this work, we further extend this idea by showing how to condition such models to avoid whole classes of \emph{semantic} errors, namely queries with runtime errors and queries that generate no results.

\begin{figure}[t]
\centering
\includegraphics[width=\linewidth]{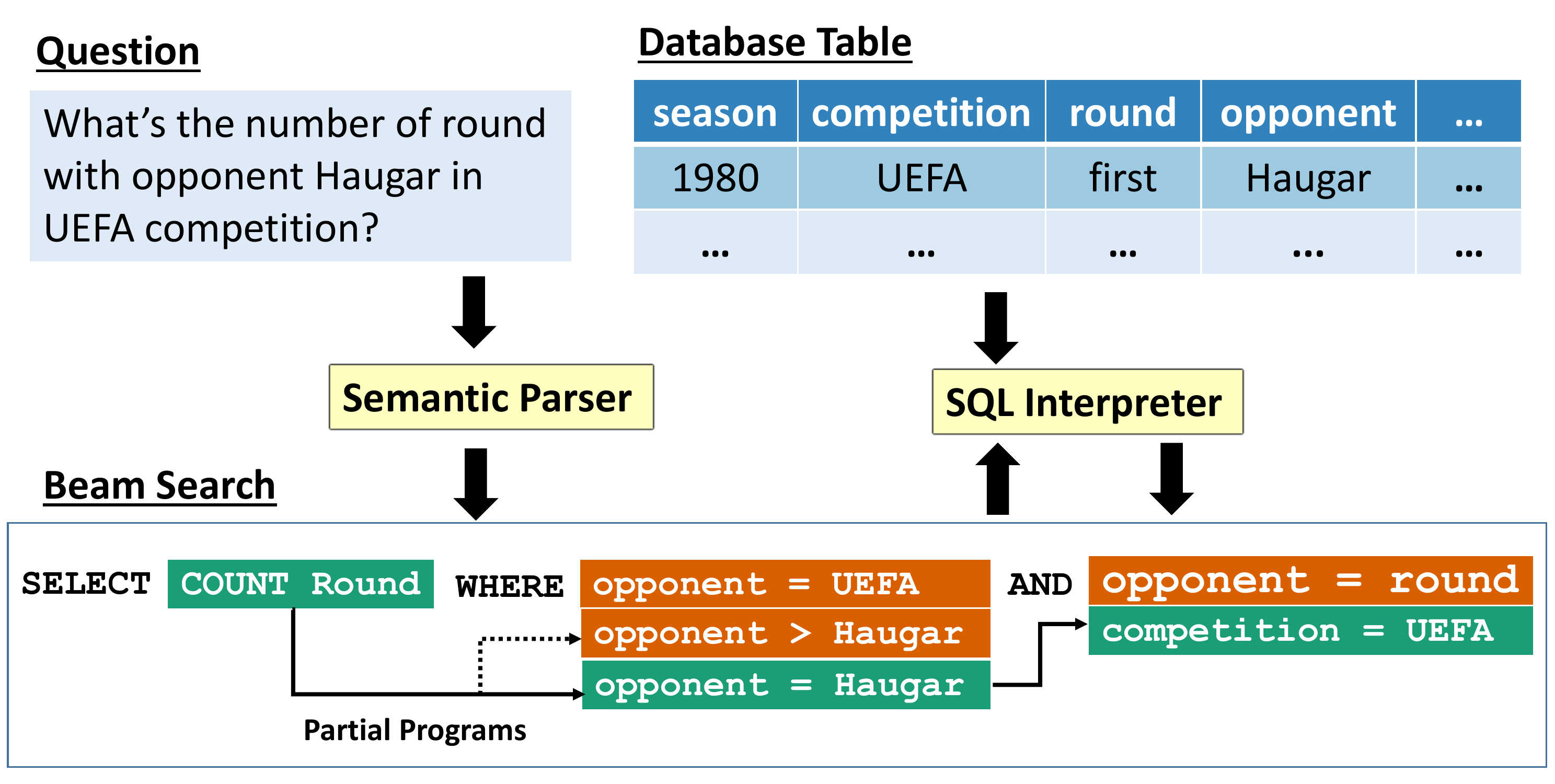}
\caption{An execution-guided decoder evaluates partially generated queries at appropriate timesteps and then excludes those candidates that cannot be completed to a correct SQL query (red background).
Here, ``\texttt{opponent~> Haugar}'' would yield a runtime error, whereas ``\texttt{opponent~= UEFA}'' would yield an empty result.}
\label{fig:overview}
\end{figure}

The key insight is that in languages such as SQL, a partially generated query can already be executed, and the results of that execution can be used to guide the generation procedure.
We call this idea \emph{execution guidance} and illustrate it in Fig. \ref{fig:overview}.
Note that the concept can be extended beyond what we are demonstrating in this paper: whereas we only use execution of partial programs to filter out results that cannot be completed to a correct answer, a more advanced execution guidance mechanism could take partial results into account to improve decision-making, for example by considering which literals occur after filtering according to a partial query.
In other words, execution guidance extends standard autoregressive decoders to additionally condition them on non-differentiable partial execution results at appropriate timesteps.

We show the effectiveness of execution guidance by extending a range of existing models with execution guidance and evaluating the resulting models on various text-to-SQL tasks with different scale and query complexity.
Concretely, we first extend four state-of-the-art semantic parsing models with execution guidance and then demonstrate that execution guidance universally improves their performance on different text-to-SQL datasets.
The considered models cover two widespread families of text-to-SQL semantic parsers: \textbf{(a)} autoregressive generative models such as Pointer-SQL~\cite{chenglong,wang2018execution} and Seq2Seq with attention~\cite{data-atis-geography-scholar}, as well as
\textbf{(b)} template \& slot-filling based models (a baseline by~\citet{data-sql-advising} and Coarse2Fine by~\citet{coarse2fine}).
We evaluate the unmodified baselines as well as our extensions on the WikiSQL~\cite{WikiSQL}, ATIS~\cite{data-atis-original}, and GeoQuery~\cite{geoquery} datasets, showing that using the execution guidance paradigm during decoding leads to an improvement of $1\% - 6\%$ over the base models.
As a result, our extension of Coarse2Fine with execution guidance becomes the state of the art model on the WikiSQL dataset with 83.8\% execution accuracy.

\section{Execution-Guided Decoding}

\label{sec:execution}

As discussed above, the key insight in this work is that partially generated SQL queries can be executed to provide guidance for the remainder of the generation procedure.
In this work, we only use this information to filter out partial results that cannot be completed to a correct query, for example because executing them yields a runtime error or because the generated query constraints already yield no results.
We discuss these cases in detail now.

\subsubsection{Execution Errors}
We consider the following two types of errors that could be identified by the execution engine:
\begin{itemize}[itemsep=1pt,topsep=2pt]
    \item \emph{Parsing errors}: A program $p$ causes a parsing error if it is syntactically incorrect. This kind of error is more common for complex queries (as appearing in the GeoQuery and ATIS datasets). Autoregressive models are more prone to such errors than template-based and slot-filling-based models.
    \item \emph{Runtime errors}: A program $p$ throws a run-time error if it has a component whose operator type mismatches its operands types. Such an error could be caused by a mismatch between an aggregation function and its target column (e.g., sum over a column with string type) or a mismatch between a condition operator and its operands (e.g., applying $>$ to a column of float type and a constant of string type).
\end{itemize}
In these cases, the decoded program cannot be executed, and hence cannot yield a correct answer.
If we can assume that every query \emph{has} to yield a result, we consider an additional type of error:
\begin{itemize}[itemsep=1pt,topsep=2pt]
    \item \emph{Empty output}: When executed, a program $p$ could return a empty result if the predicate generated by the decoder is overly restrictive (e.g., a predicate $c = v$ is generated but the constant $v$ does not exist in the column $c$).
\end{itemize}
Note that in real-world situations it is common to expect that the query output is non-empty.
For example, data scientists and database administrators often experiment with partial queries in order to compose a desired query from subtables, and use non-empty result as an indicator of the query's correctness~\cite{IyyerSQA2016}.

\subsubsection{Using Execution Guidance in Decoding}
To avoid generating queries yielding the errors discussed above, we integrate the query generation procedure with a SQL execution component.
Extending a model with execution guidance thus requires to pick specific stages of the generative procedure at which to execute the partial result, and then use the result to refine the remaining generation procedure.

\begin{algorithm}[t]
\caption{Execution-guided decoding as an extension of a standard recurrent autoregressive decoder.}
\label{alg:egdecoder}
    \begin{algorithmic}[1]
    \Procedure{EG-Decoding}{encoded query $O$, encoded table columns \columns, beam size $k$}
        \State $\mathbf{h}_0 \gets$ an initial hidden decoder state
        \ForAll{steps $1 \le t \le T$ of $k$-beam decoding}
            \State \textit{\# Compute $k$ new decoder states in the beam:}
            \State $\mathbf{h}_t^1,\dots,\mathbf{h}_t^k \gets \Call{Decode}{O, \columns, \mathbf{h}_{t-1}^1, \dots, \mathbf{h}_{t-1}^{k}}$
            \State $P_t^1, \dots, P_t^k \gets$ partial programs corresponding to the states $\mathbf{h}_t^1, \dots, \mathbf{h}_t^k$
            \If{the current stage $t$ is executable}
                \State \textit{\# Retain only the top $k$ \textbf{executable} programs:}
                \For{$1 \le i \le k$}
                    \If{$P_t^i$ has an error or empty output}
                        \State Remove $\mathbf{h}_t^i$ from the beam
                    \EndIf
                \EndFor
            \EndIf
        \EndFor
        \State \Return{the top-scored program $\argmax\limits_{1 \le i \le k} \Pr(P_T^i)$}

\EndProcedure
\end{algorithmic}
\end{algorithm}


We show the pseudocode of an execution-guided extension of a standard autoregressive recurrent decoder in \rA{alg:egdecoder}.
It can be viewed as an extension of standard beam search applied to a model-specific decoder cell \textsc{Decode}.
Whenever possible (i.e., when the result at the current timestep $t$ corresponds to an executable partial program), the procedure retains only the top $k$ states in the beam that correspond to the partial programs without execution errors or empty outputs.

In non-autoregressive models based on feedforward networks, execution guidance can be used as a filtering step at the end of the decoding process, e.g., by dropping result programs that yield execution errors.
The same can be applied to any autoregressive decoder at the end of beam decoding.
However, in many application domains (including SQL generation), it is possible to apply execution checks to \emph{partially} decoded programs, and not just at the end of beam decoding.
For example, this allows to eliminate an incorrectly generated string-to-string inequality comparison ``\dots~\texttt{WHERE opponent~> \textquotesingle Haugar\textquotesingle}~\dots'' from the beam immediately after the token \texttt{\textquotesingle Haugar\textquotesingle} is emitted (see \rF{fig:overview})
As our experiments show, this significantly improves the effectiveness of execution guidance.
The exact frequency and stages where execution guidance can be applied depends on the underlying decoder model.


\section{Base Models}
\label{sec:model}

In this section, we describe the base models that we augmented with execution guidance and the details of our execution-guided decoder implementation for them.
We only provide high-level model descriptions and refer to the respective source papers for details.
In total, we extended four base models from prior work, which at the time of evaluation had achieved state-of-the-art performance on the WikiSQL, ATIS, or GeoQuery datasets.
Two of them~\cite{chenglong,data-atis-geography-scholar} are \emph{generative}, employing a recurrent sequence-to-sequence decoder.
The other two~\cite{coarse2fine,data-sql-advising} are \emph{slot-filling}, employing a decoder that fills in the holes in a template (which itself can be generated with a sequence-to-sequence model).
These models illustrate a wide variety of autoregressive decoders that can be augmented with execution guidance.

\subsection{Seq2Seq with Attention \small\cite{data-atis-geography-scholar}}
\citet{data-atis-geography-scholar} apply a relatively standard sequence-to-sequence model to text-to-SQL tasks.
The natural language input is encoded using a bidirectional RNN with LSTM~\cite{Hochreiter97} cells, and the target SQL query is directly predicted token by token.
Additionally, an attention mechanism~\cite{cho2014learning,luong2015effective} is employed to improve the model performance.
To handle the small number of data points in in some datasets, the encoding step uses pre-trained word embeddings from word2vec~\cite{word2vec}, which are concatenated to the embeddings that are learned for tokens from the training data.

\subsubsection{Integrating Execution Guidance}
In classical sequence-to-sequence models, it is unclear a priori at which stages of the decoding we have a valid partial program.
Thus, it is hard to implement partial program execution statically.
We thus use the simplest form of execution-guidance: we first perform standard beam decoding with width $k$, and then choose the highest-ranked generated SQL program without execution errors at the end of the decoding.

\subsection{Pointer-SQL \small\cite{chenglong}}

The Pointer-SQL model introduced by \citet{chenglong} extends and specializes the sequence-to-sequence architecture to the WikiSQL dataset.
It takes a natural language question and a table schema of a single table $t$ as inputs.
To encode the inputs, a bidirectional RNN with LSTM cells is used to process the concatenation of the table header (column names) of the queried table and the question as input to learn a joint representation.
The decoder is another RNN that can attend over and copy from the encoded input sequence.
The key characteristic of this model is that the decoder uses three separate output modules corresponding to three decoding types.
One module is used to generate SQL keywords, one module is used to copy a column name from the table header, and one module can copy literals from the natural language question.

\begin{figure}[t]
\centering
\includegraphics[width=\linewidth]{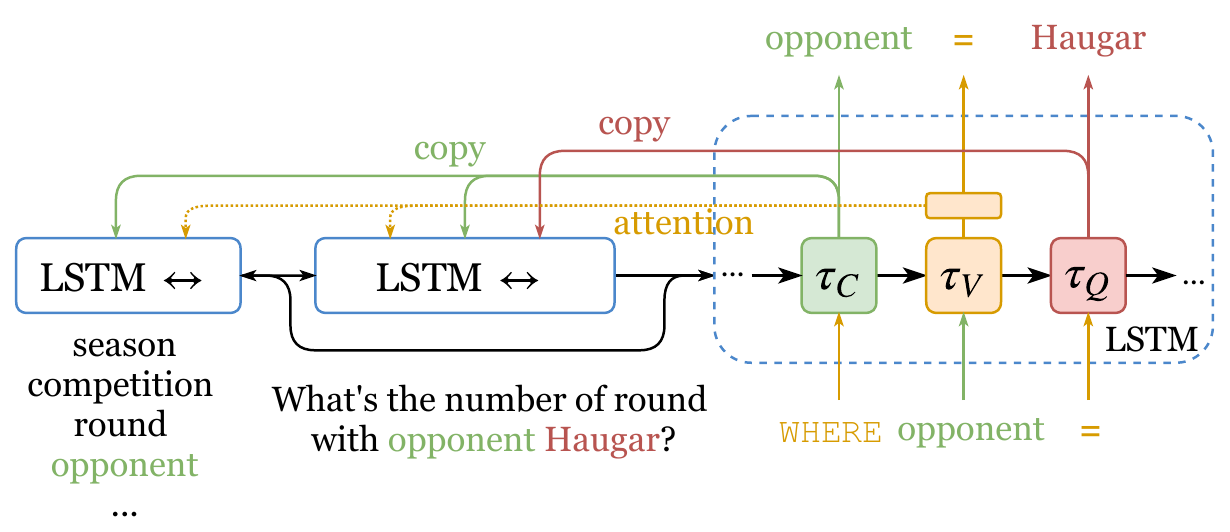}
\caption{Overview of the Pointer-SQL model. The model encodes table columns as well as the user question with a BiLSTM and then decodes the hidden state with a typed LSTM, where the decoding action for each cell is statically determined. Source:~\citet{chenglong}.}
\label{fig:pointersql}
\end{figure}

The grammar of SQL expressions in the WikiSQL dataset can be described by the regular expression ``\texttt{Select} $f$ $c$ \texttt{From} $t$ \texttt{Where} ($c$ $op$ $v$)$^*$''.
Here $f$ refers to an aggregation function, $c$ refers to a column name, $t$ refers to the table name, $op$ refers an comparator and $v$ refers to a value.
The Pointer-SQL model exploits this standardized shape of WikiSQL queries to select which output module to use at each decoding step, as shown in \rF{fig:pointersql}.

\subsubsection{Integrating Execution Guidance}
We extend the model to use execution guidance in two cases.
First, after decoding the aggregation operator $f$ and the aggregation column $c$, we run the execution engine over the partial program ``\texttt{Select $f$ $c$ Where True}'' to determine whether $f$ and $c$ are compatible.
We replace decoding results failing this check by those pairs of $f'$ and $c'$ from the set of valid operator/column pairs with the next-highest joint probability according to the token distribution produced by the decoder; and then proceed to the decoding of predicates.

Second, after decoding a predicate $c_1~\mathit{op}~c_2$, we evaluate the partial program including the predicate to check whether the predicate triggers a type error or results in an empty output.
Again, we replace decoding results failing this check by new predicates $c_1'~\mathit{op}'~c_2'$ with the next-highest joint probability from the set of error-free predicates.

In practice, instead of computing all correct choices, we parameterize the execution-guided decoder with a \emph{beam width} $k$ to restrict the number of alternative tokens considered at each decoding step and simply discard those results that trigger errors.
As described earlier, this approach resembles a standard beam decoder where instead of generating the top-$k$ results with the highest probability, we additionally use evaluation results to discard erroneous programs.

\subsection{Template-Based Model \small \cite{data-sql-advising}}
The template-based approach to text-to-SQL generation introduced as baseline by \citet{data-sql-advising} also exploits the simple structure of target queries.
First, the dataset is preprocessed to extract the most common \emph{templates} -- program sketches in which the operands in conditionals have been replaced by slots.
The template-based model makes two kinds of predictions: \textbf{(a)} which template to use, and \textbf{(b)} which words in the question should be used to fill the slots in the chosen template.
For this, a bidirectional RNN is run over the natural language question, outputting a ``used in slot'' or ``not used in query'' signal for each token.
A small fully-connected network is then used to predict the chosen template from the final states of the RNN.
The output query is constructed by filling the slots from the template with the predicted tokens from the input question; but as no agreement between the two predictions is enforced, this can easily fail.
\rF{fig:model-template-baseline} illustrates the structure of this template-based model.
As the model is restricted to templates seen only during training, it cannot generalize to query structures that do not appear in the training set.

\begin{figure}[t]
\centering
\includegraphics[width=\linewidth]{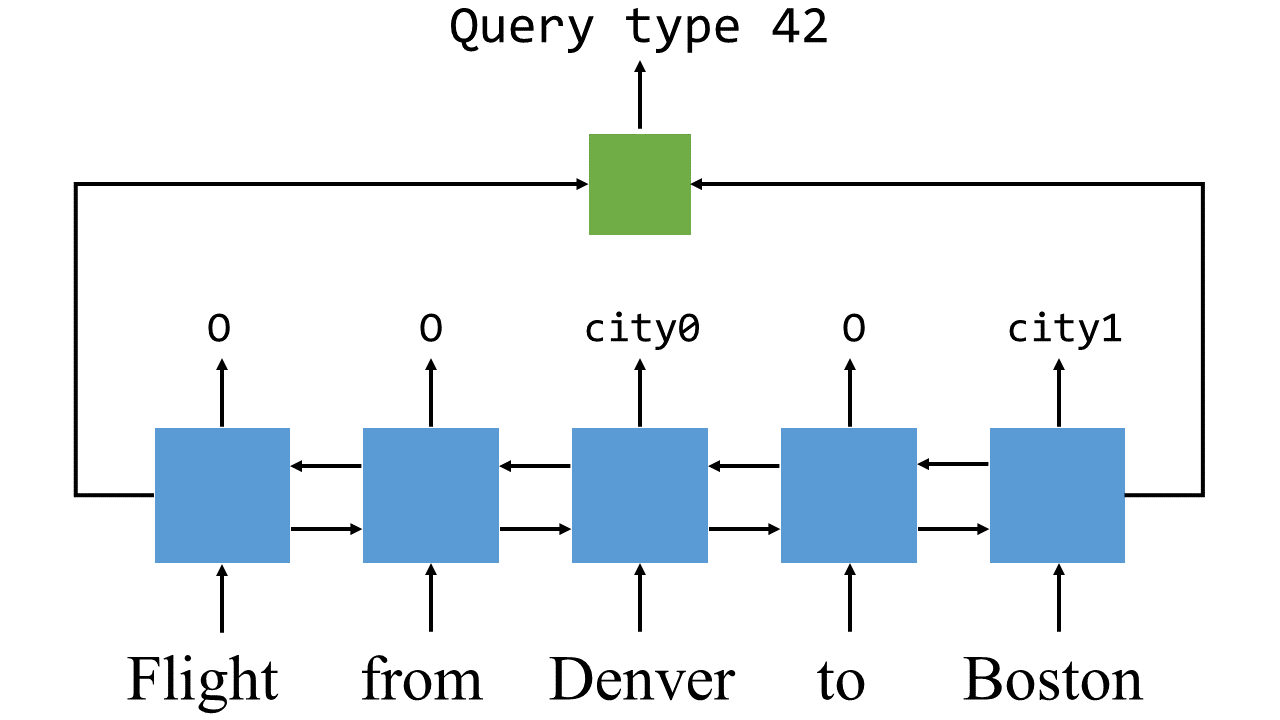}
\caption{Overview of the template-based baseline model. The encoder consists of an Bi-LSTM, whose outputs are then used by feed-forward networks to determine the template and fill in the slots. Source:~\citet{data-sql-advising}.}
\label{fig:model-template-baseline}
\end{figure}

\subsubsection{Integrating Execution Guidance}
For a given beam width of $k$, we independently pick the top-$k$ choices for the template and for the slots using beam decoding. The final top-$k$ template-slot candidate predictions are decided based on the joint probability of the template and the slots. We then pick the SQL program with the highest joint probability which does not lead to an execution error. Partial program execution is not possible in this model, as the model does not employ an autoregressive recurrent decoder.

\subsection{Coarse2Fine \small\cite{coarse2fine}}
The Coarse2Fine model can be viewed as a mix of template-based models and sequence-to-sequence models.
It is a two-stage model for general text-to-code translation, where the first stage generates a coarse ``sketch'' (a template) of the target program and the second stage fills its missing ``slots''.

The model generates programs in the following three steps. First, the input question (and in the case of WikiSQL, the table schema) is encoded using a bidirectional RNN using LSTM cells. Then, the \emph{sketch generator} uses a classifier to choose a query sketch of the form ``\texttt{Where} ($op$)$^*$'' from one of the predefined sketches. Intuitively, the sketch determines the number of conditions in the \texttt{Where}-clause as well as the comparison operators. Finally, the \emph{fine meaning decoder} uses the inputs as well as the generated sketch to produce the full query by filling in slots. Similarly to Pointer-SQL, the fine-decoding model uses a copying mechanism to generate column names and constants.




\subsubsection{Integrating Execution Guidance}
We cannot apply execution-guidance to the sketch generator of the Coarse2Fine model, as the generated sketch is not an executable partial program. Here, we simply pick the most probable sketch, and proceed to the next stage of decoding.

Similarly to the execution-guidance implementation for the Pointer-SQL model, after decoding the aggregation operator $f$ and the aggregation column $c$ we run the execution engine over the partial program ``\texttt{Select $f$ $c$ From $t$ Where True}'', and pick a \emph{compatible} $(f, c)$ pair with the highest joint probability.

In the slot-filling stage, when \texttt{Where}~$c_1~op~c_2$ templates are completed using the fine meaning decoder, we apply execution guidance in a similar way as in the Pointer-SQL model. As the $op$ operations were previously generated as part of the sketch, we retain only the $k$ highest-ranked $(c_1, c_2)$ combinations that do not result in an execution error. If no valid choices are found, we backtrack and emit a different sketch from the ``coarse'' model.




\section{Experimental Results}
\label{sec:experiments}

\begin{figure}
\centering
\includegraphics[width=\linewidth]{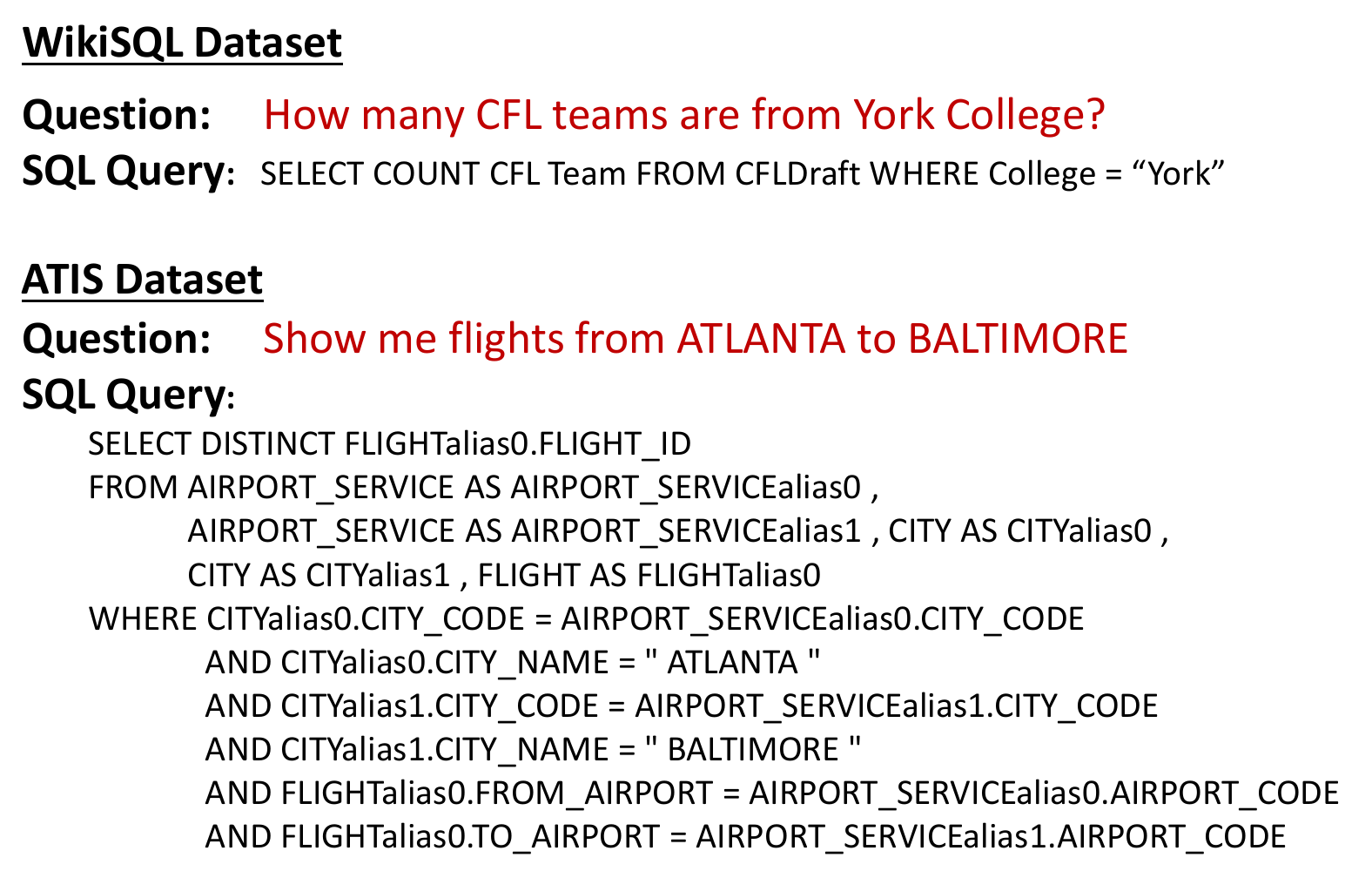}
\caption{Example of WikiSQL and ATIS queries}
\label{fig:atis}
\end{figure}

We evaluate the effect of execution guidance by comparing the baseline models with our new variants that integrate execution guidance.
Because this guidance only requires changes to the generation procedure at test time, we can use the same pre-trained models in all instances.
As discussed above, our four different models cover two different families (generative and slot-filling) and integrate execution guidance to different degrees.

For each dataset, we pick a model from each family that has achieved state-of-the-art performance at the time of evaluation.
Thus, specifically, we evaluate the Pointer-SQL model~\cite{chenglong} and the Coarse2Fine~\cite{coarse2fine} model on the WikiSQL dataset. The template-based baseline~\cite{data-sql-advising} and the Seq2Seq with Attention~\cite{data-atis-geography-scholar} models are evaluated on the ATIS and GeoQuery datasets.

\subsection{WikiSQL Experiments}
\label{sec:exp-setup}
WikiSQL~\cite{WikiSQL} is a recently introduced natural language to SQL dataset, consisting of 80,654 pairs of questions and SQL queries distributed across 24,241 tables from Wikipedia.
The task is to generate the correct SQL query for a given natural language question and a table schema (i.e., table column names), without using the content values of tables.
The SQL structure in the WikiSQL dataset is simple and always follows the structure \texttt{SELECT agg sel WHERE (col op cond)*}.
Also, the natural language question is often grammatically wrong (see \rF{fig:overview} for an example).
Nonetheless, the WikiSQL dataset poses a good challenge for text-to-SQL systems and has seen significant interest since its release~\cite{coarse2fine,wang2018execution,huang2018ptmaml,DBLP:conf/naacl/YuLZZR18}.
In our experiments, we use the default train/development/test split, leading to 56,324 training pairs, 8,421 development pairs, and 15,878 test pairs.
Each table is present in only one split to test generalization to unseen tables.

\begin{table}[t]
\centering
\small
\begin{tabular}{lcccc}
\toprule
\multirow{ 2}{*}{Model}& \multicolumn{2}{c}{Dev} & \multicolumn{2}{c}{Test} \\ \cmidrule{2-5}
  & Acc$_\text{syn}$    & Acc$_\text{ex}$ & Acc$_\text{syn}$    & Acc$_\text{ex}$ \\
  \midrule
Pointer-SQL (\citeyear{chenglong}) & 61.8 & 72.5 & 62.3 & 71.9 \\
Pointer-SQL  + EG (3) & 66.6 & 77.3 & 66.7 & 76.9\\
Pointer-SQL  + EG (5) & \textbf{67.5} & \textbf{78.4} & \textbf{67.9} & \textbf{78.3}\\ \midrule
Coarse2Fine (\citeyear{coarse2fine}) & 72.9 & 79.2 & 71.7 & 78.4\\
Coarse2Fine + EG (3)  & 75.6 & 83.4 & 74.8 & 83.0\\
Coarse2Fine + EG (5)  & \textbf{76.0} & \textbf{84.0} & \textbf{75.4} & \textbf{83.8} \\
\bottomrule
\end{tabular}
\caption{Test and Dev accuracy (\%) of the models on WikiSQL data, where Acc$_\text{syn}$ refers to syntactical accuracy and Acc$_\text{ex}$ refers to execution accuracy. ``+ EG ($k$)'' indicates that model outputs are generated using the execution-guided strategy with beam size $k$.}
\label{result-table}
\end{table}

\begin{table*}[t]
\centering
\begin{tabular}{lcccc}
 \toprule
  \multirow{ 2}{*}{Model}                           & \multicolumn{2}{c}{ATIS}                & \multicolumn{2}{c}{GeoQuery}\\
 \cmidrule{2-5}
                                                    & Dev Acc$_\text{ex}$ & Test Acc$_\text{ex}$ & Dev Acc$_\text{ex}$ & Test Acc$_\text{ex}$\\
 \midrule
  Template-based (\citeyear{data-sql-advising})     & 35.1                & 32.6                & 50.5               & 55.2\\
  Template-based + EG (5)                           & 36.7                & 37.1                & \textbf{52.7}               & 55.2\\
  Template-based + EG (10)                          & \textbf{37.6}       & \textbf{37.6}                & \textbf{52.7}               & \textbf{55.6}\\
 \midrule
  Seq2Seq (\citeyear{data-atis-geography-scholar})  & 78.6                & 77.0                & 76.0               & 72.5\\
  Seq2Seq + EG (5)                                  & \textbf{78.8}                & 77.3                & \textbf{78.0}               & \textbf{75.0}  \\
  Seq2Seq + EG (10)                                 & \textbf{78.8}                & \textbf{77.9}                & \textbf{78.0}               & \textbf{75.0} \\
 \bottomrule
\end{tabular}
\caption{{Test and Dev accuracy (\%) of the models on ATIS and GeoQuery data, where Acc$_\text{ex}$ refers to execution accuracy. ``+~EG~($k$)'' indicates that model outputs are generated using the execution guiding strategy with beam size $k$.\footnotemark}}
\label{result-table-2}
\end{table*}

\rTab{result-table} shows the results for Pointer-SQL and Coarse2Fine.
We report both the syntactical accuracy Acc$_\text{syn}$ corresponding to the ratio of predictions that are exactly the ground truth SQL query, as well as the execution accuracy Acc$_\text{ex}$ corresponding to the ratio of predictions that return the same result as the ground truth when executed. Note that the execution accuracy is higher than syntactical accuracy as syntactically different programs can generate the same results (e.g., programs differing only in predicate order).
In execution-guided decoding, we report two model variants, one using a beam size of $3$ and the other a beam size of $5$.

The comparison results show that execution-guided decoding significantly improves both syntactical accuracies as well as execution accuracies of the models. Using a beam size of 5 leads to an improvement of 6.4\% (71.9\% to 78.3\%) for the Pointer-SQL model, and an improvement of 5.4\% (78.4\% to 83.8\%) for the Coarse2Fine model on the Test dataset (see \rF{fig:examples-EG} for some examples).
Similar improvements are observed on the Dev dataset.
To the best of our knowledge, this improvement on the Coarse2Fine model makes it the new state of the art in terms of execution accuracy on WikiSQL.

\begin{figure}
\centering
\includegraphics[width=\linewidth]{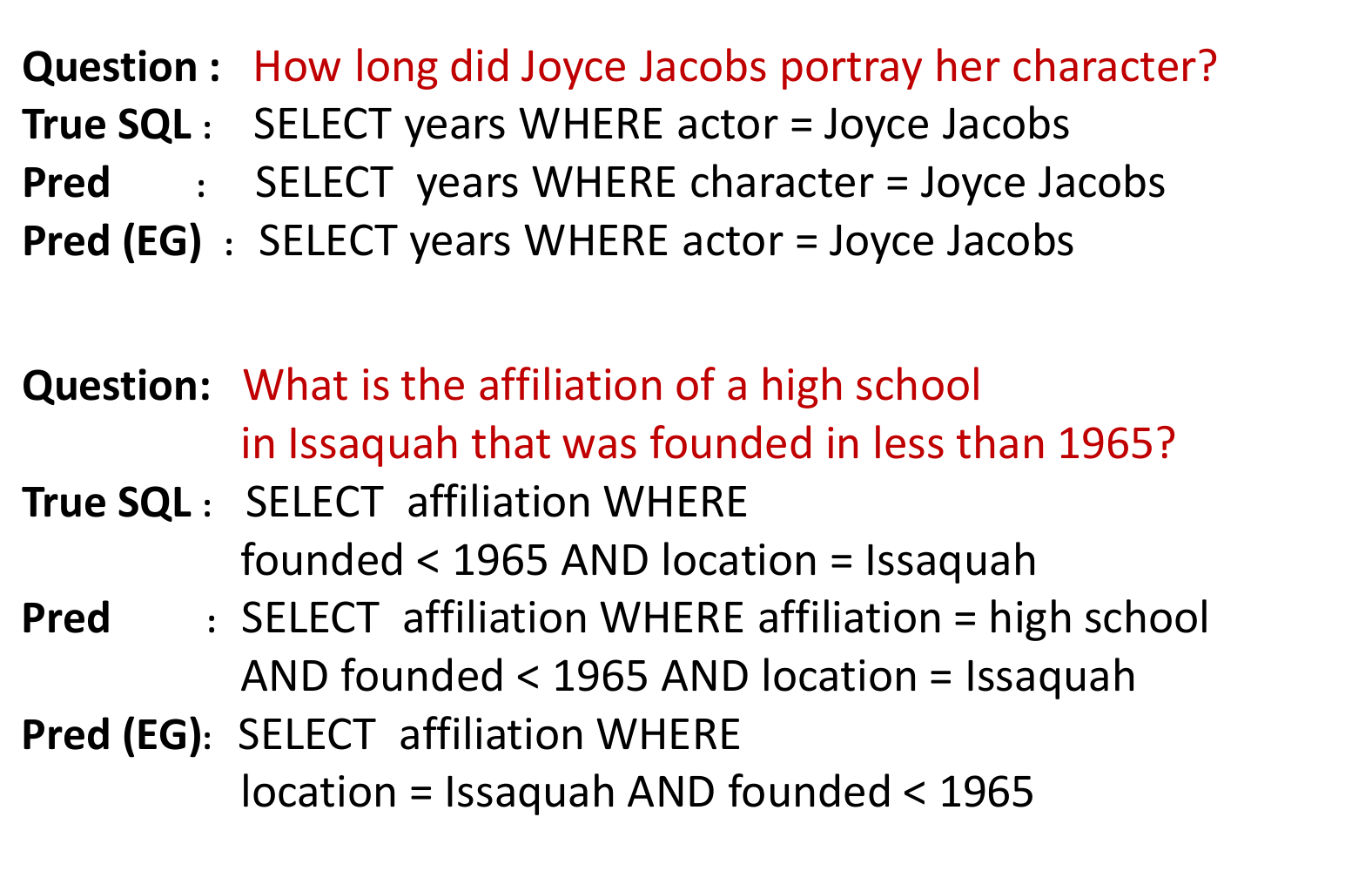}
\caption{Some examples where execution guidance (EG) for Coarse2Fine leads to correct prediction. In the first example, the table column is corrected by execution guidance due to an empty output. 
In the second example, the execution guidance corrects the sketch as all possible slot-filling options for the three-condition sketch overconstrain the program and also yield an empty output.
The experiments were performed with beam size of 5. }
\label{fig:examples-EG}
\end{figure}

\subsection{ATIS and GeoQuery Experiments}

We use the ATIS and GeoQuery datasets from the standardized collection of datasets~\cite[ver. 1]{data-sql-advising} on the default train/test/dev split. Compared to the WikiSQL data, these datasets consider substantially more complex SQL queries, for example requiring to access multiple tables in one SQL query. The ATIS dataset on average references 6 tables per query, while the GeoQuery dataset on average references 3 tables per query. There are also other complexities in the SQL programs, such as nesting, inner joins, and group-by operations.
See \rF{fig:atis} for an example.

\rTab{result-table-2} shows the results for the template-based model~\cite{data-sql-advising} and the sequence-to-sequence model~\cite{data-atis-geography-scholar} for the ATIS and GeoQuery datasets. We compare the performance of the base models with execution guided decoding with beam size 5 and 10. We report execution accuracy for evaluation on both the dev and test splits of the data.

The comparison result shows that execution-guided decoding leads to an improvement ranging from 0.4\% to 5\% on the test data. The template-based model improves most by 5\% on the ATIS dataset, and by 0.4\% on the GeoQuery dataset. The significantly smaller improvement can be attributed to the slot-filling nature of the model (which is less amenable to partial execution guidance) along with the simpler structure of the GeoQuery dataset. Conversely, the sequence-to-sequence model sees an improvement of 2.5\% on the GeoQuery dataset and only a 0.9\% improvement on the ATIS dataset.


\subsection{Discussion \& Analysis}

\begin{table}[t]
\centering
\begin{tabular}{lcc}
\toprule
Model                                & Acc$_\text{ex}$ & Execution Errors \\
\midrule
Coarse2Fine (\citeyear{coarse2fine}) & 78.4           & 10.70 \\
Coarse2Fine + EG (3)                 & 83.0           & \phantom{0}0.03\\
Coarse2Fine + EG (5)                 & 83.8           & \phantom{0}0.01\\
\bottomrule
\end{tabular}
\caption{Accuracy (\%) and the overall amount of execution errors (\%) on the WikiSQL test dataset.}
\label{result-table-coarse2fine-exerrors}
\end{table}

\begin{table}[t]
\centering
\begin{tabular}{lcc}
\toprule
Model                                & Acc$_\text{ex}$ & Execution Errors \\
\midrule
Seq2Seq (2017)                       & 77.0           & 10.5\\
Seq2Seq + EG (5)                     & 77.3           & \phantom{1}6.9\\
Seq2Seq + EG (10)                    & 77.9           & \phantom{1}6.3\\
\bottomrule
\end{tabular}
\caption{Accuracy (\%) and the overall amount of execution errors (\%) on the ATIS test dataset.}
\label{result-table-seq2seq-exerrors}
\end{table}

\begin{table}[t]
\centering
\begin{tabular}{lc}
\toprule
Model                                    & Acc$_\text{ex}$ (\%)   \\
\midrule
Coarse2Fine (\citeyear{coarse2fine})     & 78.4 \\
Coarse2Fine + EG (5)                     & \textbf{83.8} \\
\qquad No Aggregation execution          & 83.7 \\
\qquad No Condition execution           & 80.2 \\
\qquad No Sketch backtracking            & 82.6 \\
\bottomrule
\end{tabular}
\caption{Ablation study on different execution-guided decoding steps on the WikiSQL dataset.}
\label{result-table-coarse2fine-ablations}
\end{table}

Our experiments show that execution guidance is a simple but effective tool to improve a wide range of existing text-to-SQL models.
However, we note that this strategy simply improves the number of semantically meaningful programs and not only the number of semantically correct programs.
For this, consider Tables~\ref{result-table-coarse2fine-exerrors}~and~\ref{result-table-seq2seq-exerrors}, in which we show the number of execution errors remaining after adding execution guidance to the models.
For the Coarse2Fine model on the WikiSQL data, we see a drop of 10\% in execution errors, but only an improvement of execution accuracy of 5.4\%.
Thus, about half of the incorrect programs were replaced by correct predictions.
On the other hand, the Seq2Seq model on the ATIS data sees a drop of execution errors of only 4.2\% (the large number of remaining errors is due to the more complex nature of ATIS SQL queries).
More significantly, this only yields an improvement in accuracy of 0.9\%.
Note that our integration of execution guidance for this model is only post-hoc filtering, and thus we speculate that more fine-grained execution checks on partial programs is more effective.

\paragraph{Ablations}
To better understand the source of improvement from execution guidance, we performed additional ablation experiments on the Coarse2Fine model.
Recall that the execution guidance on the Coarse2Fine model applies at three different intermediate steps of the decoding process.
When the model generates a WikiSQL query of form ``\texttt{Select} $f$ $c$ \texttt{From} $t$ \texttt{Where} ($c$ $op$ $v$)$^*$'', the execution-guided decoder
\textbf{(a)}~checks the choice of the aggregation $f$~$c$,
\textbf{(b)}~checks the choice of each generated condition $c~op~v$, and
\textbf{(c)}~backtracks to a different sketch from the ``coarse'' stage of the Coarse2Fine model if no generated candidate programs execute correctly.
We perform ablation experiments in which we turn off execution guidance at each of these steps respectively.
We use beam size of $5$ for all the ablation experiments.

The results are shown in Table~\ref{result-table-coarse2fine-ablations} and show that execution guidance has little effect on the choice of aggregation functions, but contributes significantly in the generation of conditions.
This experiment further emphasizes the importance of fine-grained guidance that eliminates incorrect partial candidates at intermediate decoding steps.

\footnotetext{The execution accuracy results of the template-based model are lower than originally reported by \citet{data-sql-advising}, because at the time of writing, a bug was discovered in the evaluation of the original model. \citeauthor{data-sql-advising} later made changes to their template-based model. We used their published pre-trained models for our experiments.}



\section{Related Work}

\paragraph{Semantic Parsing}
Semantic parsing has been studied extensively by the natural language processing community. It maps natural language to a logical form representing its meaning, which is then used for question answering~\cite{wong:acl07,Zettlemoyer:2005:LMS,Zettlemoyer07onlinelearning,berant2013freebase},
robot navigation~\cite{chen:aaai11,tellex11}, and many other tasks. \citet{liang2016executable} surveys different statistical semantic parsers and categorizes them under a unified framework.

Recently there has been an increasing interest in applying deep learning models to semantic parsing, due to the huge success of such models on machine translation and other problems. A significant amount of work is dedicated to constraining model outputs to guarantee valid parsing results. \citet{DBLP_conf/acl/DongL16} propose the Seq2Tree model, which always generates syntactically valid trees. The same authors later propose a two-step decoding approach which first generates a rough sketch, and later uses it to constrain the final output during the second decoding pass~\cite{coarse2fine}. In contrast to these token-based decoding approaches, the work of \citet{DBLP_journals/corr/YinN17} and \citet{krishnamurthy2017neural} employ grammar-based decoding which utilizes grammar production rules as a decoding constraint.

The use of SQL as the meaning representation for semantic parsing was recently re-popularized by the introduction of the WikiSQL dataset~\cite{WikiSQL}.
A large number of subsequent neural semantic parsing works included an evaluation on WikiSQL~\cite{sqlnet,huang2018ptmaml,DBLP:conf/naacl/YuLZZR18,wang2018execution,coarse2fine}. The large number of annotated pairs of natural language queries together with their corresponding SQL representations makes it an attractive dataset for training neural network models. However, WikiSQL is often criticized for its weak coverage of SQL syntax.
\citet{data-sql-advising} later standardized a range of semantic parsing datasets with SQL as the logical representation, which supports more realistic usage of language including table joins and nested queries.
We use their standardization of the ATIS and GeoQuery datasets in our experiments.


\paragraph{Sequence-Level Objectives}
Several recently introduced techniques use reinforcement learning to incorporate sequence-level objectives into training, such as BLEU in coherent text generation \cite{Bosselut2018}, semantic coherence in dialogue generation \cite{li2016deep}, and SPIDEr in image captioning \cite{liu2017improved}.
Similarly, \citet{WikiSQL} use reinforcement learning to learn a policy with the objective of maximizing the expected correctness of the execution of generated programs.
\citet{wiseman2016sequence} propose incorporating sequence-level cost functions (such as BLEU) into beam search for sequence-to-sequence training.
These efforts are focused on integrating additional (discrete) objectives into the training objective, whereas our work only requires changes during inference time and can thus easily be applied to a wide range of different models, as shown in our experiments.


\paragraph{Program Synthesis}
The related research area of 
program synthesis aims to generate programs given some specification of user intent such as input-output examples or natural language descriptions~\cite{synthesis-survey}.
The latest work in this field combines neural and symbolic techniques in order to generate programs that are both likely to satisfy the specification and semantically correct.
This line of work includes DeepCoder~\cite{deepcoder}, Neuro-symbolic synthesis~\cite{nsps}, Neo~\cite{neo}, and NGDS~\cite{Vijayakumar18}.
They use probabilistic models to guide symbolic program search to maximize the likelihood of generating a program suitable for the specification.
Our execution guidance idea is similar to neuro-symbolic program synthesis, but in contrast, it guides a neural program generation model using a symbolic component (namely, partial program execution) to generate semantically meaningful programs.



\section{Conclusion}


For the task of SQL generation from natural language, we presented the idea of guiding the generation procedure by partial execution results.
This approach allows conditioning an arbitrary autoregressive decoder on non-differentiable partial execution results at inference time, leading to elimination of semantically invalid programs from the candidates.
We showed the widespread practical utility of execution guidance by applying it on four different state-of-the-art models which we then evaluated on three different datasets.
The improvement offered by execution-guided decoding is dependent on the nature of the extended model and the degree of integration with the decoding procedure.
Extending the Coarse2Fine model with an execution-guided decoder improves its accuracy by 5.4\% (from 78.4\% to 83.8\%) on the WikiSQL test dataset, making it the new state of the art on this task.

The idea of execution guidance can potentially be applied to other tasks, such as natural language to logical form, which we plan to explore in future work.
Furthermore, we note that in this work, we only used execution guidance to filter out incorrect predictions during generation at inference time.
We believe that integrating execution guidance into the training phase, for example by learning to make decisions conditional upon the results of executing the partial program generated so far, will further improve model performance.

\bibliographystyle{aaai}
\bibliography{nl2prog}


\end{document}